%% file: 0_Main_ICRA.tex
\documentclass[letterpaper, 10 pt, conference]{ieeeconf}  

\IEEEoverridecommandlockouts                              

\makeatletter
\let\NAT@parse\undefined
\makeatother
\usepackage[pagebackref=false,breaklinks=true,colorlinks=true,bookmarks=false,linkcolor={red!50!black},urlcolor={blue},citecolor={green!60!black}]{hyperref} 
\usepackage{url}
\usepackage{pifont}
\usepackage{booktabs} 

\usepackage{etoolbox}
\makeatletter
\patchcmd{\@makecaption}
  {\scshape}
  {}
  {}
  {}
\makeatletter
\patchcmd{\@makecaption}
  {\\}
  {.\ }
  {}
  {}
\makeatother

\usepackage{graphicx}
\usepackage{subfigure}
\usepackage{lipsum}
\usepackage{cite}
\usepackage{bm}
\usepackage{esvect}
\usepackage{amsmath}
\usepackage[ruled,vlined]{algorithm2e}
\usepackage{color}
\usepackage{amsmath,amssymb}
\usepackage{tabularx}
\usepackage{multirow}
\usepackage{arydshln}
\usepackage[dvipsnames]{xcolor}
\input{preamble}

\makeatletter
\usepackage{xspace}
\DeclareRobustCommand\onedot{\futurelet\@let@token\@onedot}
\def\@onedot{\ifx\@let@token.\else.\null\fi\xspace}

\makeatother

\definecolor{gold}{rgb}{0.82, 0.53, 0.04}
\definecolor{silver}{rgb}{0.43, 0.47, 0.35}
\definecolor{bronze}{rgb}{0.5, 0.51, 0.78}

\title{\LARGE \bf
Large-Scale Gaussian Splatting SLAM
}

\author{Zhe Xin$^{1, *}$, Chenyang Wu$^{1,2, *}$, Penghui Huang$^{1}$, Yanyong Zhang$^{2}$,  Yinian Mao$^{1}$,
Guoquan Huang$^{1,3}$ 
\thanks{$^{1}$ Meituan UAV, Beijing, China.
        {\tt\small \{xinzhe, huangpenghui03, maoyinian\}@meituan.com}}%
\thanks{$^{2}$School of Computer Science and Technology, University of Science and Technology of China, Hefei,  China.
        {\tt\small cywm39@mail.ustc.edu.cn, yanyongz@ustc.edu.cn}}%
\thanks{$^{3}$Dept. of Mechanical Engineering, Computer and Information Sciences, University of Delaware, Newark, DE, USA. {\tt\small ghuang@udel.edu}}%
\thanks{\tt\footnotesize$*$Authors contributed equally to this work.}
}

\begin{document}

\maketitle
\thispagestyle{empty}
\pagestyle{empty}

\begin{abstract}
The recently developed Neural Radiance Fields (NeRF) and 3D Gaussian Splatting (3DGS) have shown encouraging and impressive results for visual SLAM. However, most representative methods require RGBD sensors and are only available for indoor environments. The robustness of reconstruction in large-scale outdoor scenarios remains unexplored. This paper introduces a large-scale 3DGS-based visual SLAM with stereo cameras, termed {\em LSG-SLAM}. 
%
The proposed LSG-SLAM employs a multi-modality strategy to estimate prior poses under large view changes. 
In tracking, we introduce feature-alignment warping constraints to alleviate the adverse effects of appearance similarity in rendering losses. 
For the scalability of large-scale scenarios, we introduce continuous Gaussian Splatting submaps to tackle unbounded scenes with limited memory. Loops are detected between GS submaps by place recognition and the relative pose between looped keyframes is optimized utilizing rendering and feature warping losses. 
After the global optimization of camera poses and Gaussian points, a structure refinement module enhances the reconstruction quality.
With extensive evaluations on the EuRoc and KITTI datasets, LSG-SLAM achieves superior performance over existing Neural, 3DGS-based, and even traditional approaches. 
Project page: \href{https://lsgsslam.github.io}{\textit{https://lsg-slam.github.io}}.

\end{abstract}

\input{1_Introduction}
\input{2_RelatedWork}
\input{3_Method}
\input{4_Experiments}
\input{5_Conclusion}

\footnotesize


\bibliographystyle{IEEEtran} 
\bibliography{icra_abrv}

\end{document}

%% file: preamble.tex
\usepackage{overpic}
\usepackage{overpic} 
\usepackage{color}
\usepackage{mathtools} 
\usepackage{currfile}
\usepackage{multirow}

\definecolor{turquoise}{cmyk}{0.65,0,0.1,0.3}
\definecolor{purple}{rgb}{0.65,0,0.65}
\definecolor{dark_green}{rgb}{0, 0.5, 0}
\definecolor{orange}{rgb}{0.8, 0.6, 0.2}
\definecolor{red}{rgb}{0.8, 0.2, 0.2}
\definecolor{darkred}{rgb}{0.6, 0.1, 0.05}
\definecolor{blueish}{rgb}{0.0, 0.3, .6}
\definecolor{light_gray}{rgb}{0.7, 0.7, .7}
\definecolor{pink}{rgb}{1, 0, 1}
\definecolor{greyblue}{rgb}{0.25, 0.25, 1}

\renewcommand{\paragraph}[1]{\vspace{.5em}\noindent\textbf{#1}.}

\usepackage{lipsum}

\usepackage{blindtext}

\newcommand{\kostas}[1]{{\color{Bittersweet} {[\bf Kosta: #1]}}}
\newcommand{\andrew}[1]{{\color{BlueViolet} {[Andrew: #1]}}}
\newcommand{\vitto}[1]{{\color{OrangeRed} {[Vitto: #1]}}}
\newcommand{\JB}[1]{{\color{OliveGreen} {[Jon: #1]}}}
\newcommand{\tom}[1]{{\color{RoyalPurple} {[Tom: #1]}}}
\newcommand{\pratul}[1]{{\color{Emerald} {[Pratul: #1]}}}
\newcommand{\at}[1]{{\color{blueish}#1}}
\newcommand{\AT}[1]{{\color{blueish}{\bf [Andrea: #1]}}}
\newcommand{\At}[1]{\marginpar{\tiny{\textcolor{blueish}{#1}}}}
\newcommand{\al}[1]{\textbf{\color{orange}[AL: #1]}}
\renewcommand{\kostas}[1]{}
\renewcommand{\andrew}[1]{}
\renewcommand{\vitto}[1]{}
\renewcommand{\JB}[1]{}
\renewcommand{\tom}[1]{}
\renewcommand{\pratul}[1]{}
\renewcommand{\at}[1]{}
\renewcommand{\AT}[1]{}
\renewcommand{\At}[1]{}
\renewcommand{\al}[1]{}


%% file: 1_Introduction.tex
\section{Introduction}
Visual SLAM is an enabling technology for spatial intelligence in autonomous robotics and embodied AI.
From the perspective of map representation, SLAM can be categorized into sparse, dense, implicit neural representation, and explicit volumetric representation.
Traditional sparse~\cite{davison2007monoslam, klein2007parallel, campos2021orb, engel2014lsd, forster2014svo, engel2017direct} and dense~\cite{newcombe2011kinectfusion, dai2017bundlefusion, whelan2015elasticfusion} SLAM systems emphasize geometric mapping and heavily rely on handcrafted features. Moreover, these methods only reliably account for the observed portions of the scene during the reconstruction (mapping).
Implicit neural representation, particularly Neural Radiance Fields (NeRF)~\cite{mildenhall2021nerf} is learned via differentiable rendering and can create high-quality, novel camera viewpoints. However, per-pixel ray marching remains a significant bottleneck for rendering speed~\cite{zhu2023nicer}. Moreover, the implicit features are embedded by multi-layer perceptrons (MLP), which may suffer from catastrophic forgetting~\cite{zhu2022nice} and the implicit scene is not easy to edit~\cite{chen2024gaussianeditor}. 

On the other hand, 3D Gaussian Splatting (3DGS)~\cite{kerbl20233d} explicitly represents a scene using Gaussian points, the rasterization of 3D primitives allows 3DGS to capture high-fidelity 3D scenes while facilitating faster rendering. 
SplaTAM~\cite{keetha2024splatam} enhances the rendering quality by eliminating view-dependent appearance and employing isotropic Gaussian points. MonoGS~\cite{matsuki2024gaussian} adopts a map-centric approach, dynamically allocating Gaussian points to enable the modeling of arbitrary spatial distributions. However, these methods use a simple uniform motion model to predict prior poses, which is easy to drift with significant viewpoint changes. Additionally, they lack an explicit loop closure module to eliminate accumulated errors and are only tested in small-scale indoor environments.



\begin{figure}[t]
    \centering
    \includegraphics[width=0.85\linewidth]{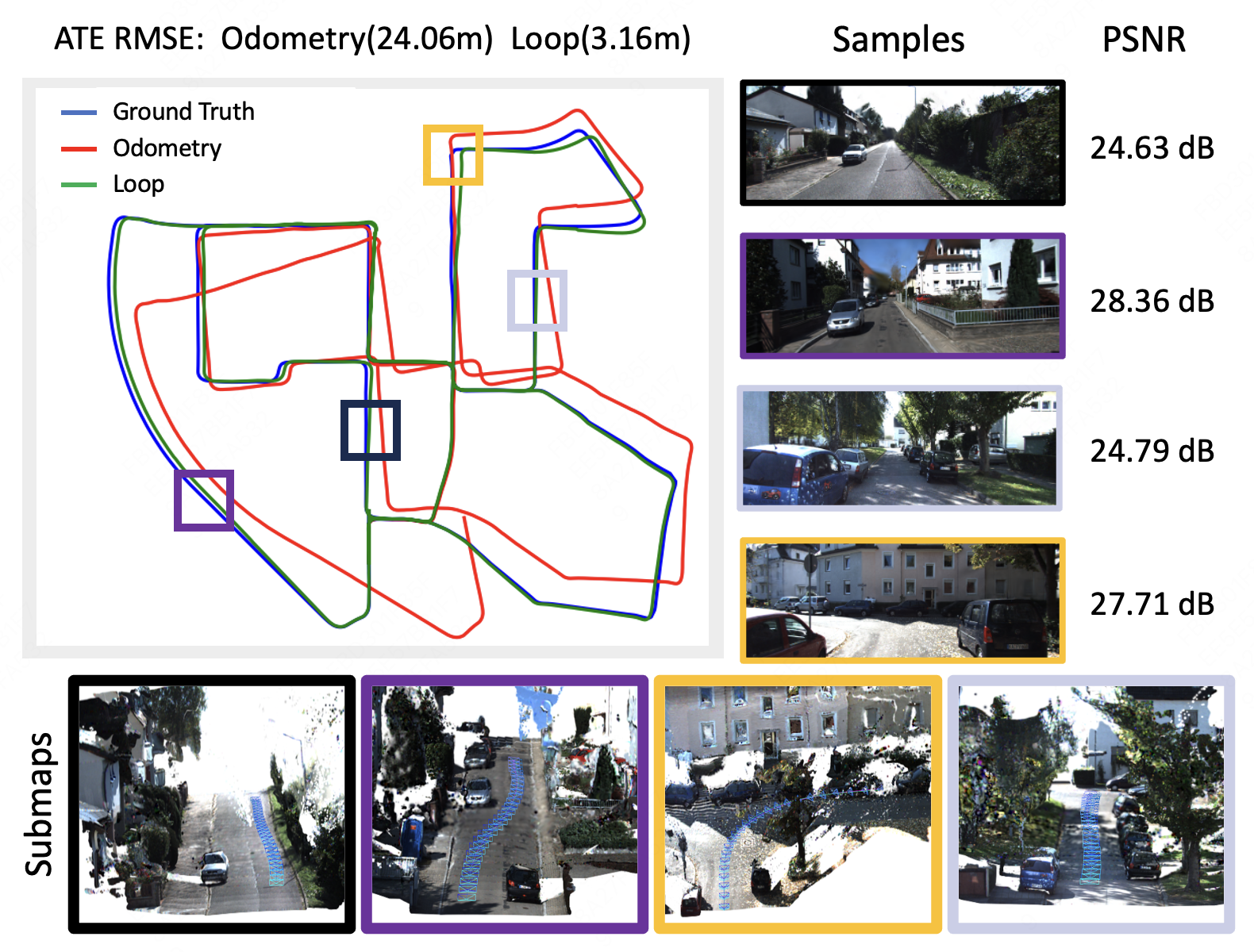}
    \caption{Results on KITTI sequence 00. LSG-SLAM enables precise camera tracking and high-fidelity reconstruction in challenging large-scale scenarios. We showcase the trajectory error before and after loop closure and some sample keyframes with associated submaps. The PSNR is evaluated after the structure refinement.
    }
    \label{fig:intro}
    \vspace{-5mm}
\end{figure}

In this paper, we develop an efficient large-scale 3DGS-based stereo visual SLAM system, termed {\em LSG-SLAM}, 
the first of its kind 3DGS-SLAM specifically designed for large-scale (outdoor) environments 
(e.g., see Fig.~\ref{fig:intro}). 
In particular, we adopt a multi-modality strategy for pose tracking to address large view changes between frames. 
In pose optimization, we integrate rendering losses and feature-alignment warping constraints. The former helps mitigate detection and matching errors caused by feature points, while the latter alleviates the adverse effects of appearance similarity. These improvements enable our method to operate at low frame rates, catering to data-limited situations. 
For mapping in large-scale scenarios, we introduce continuous GS submaps to handle scalability issues caused by unbounded scenes with limited memory. 
Valid loops are carefully detected by place recognition between keyframes in different GS submaps.
Leveraging the rasterization of Gaussian Splatting, loop constraints are estimated by minimizing the difference between the rendered and the query keyframes, using the same losses as tracking. Furthermore, a submap-based structure refinement module enhances the reconstruction quality after the global pose graph and point cloud adjustment.


In summary, the contributions of our work include:
\begin{itemize}
    \item We develop the first of its kind system of 3DGS-based stereo visual  SLAM  for large-scale environments, which significantly improves tracking stability, mapping consistency, scalability, and reconstruction quality.
    \item We advocate efficient 3DGS rendering to generate novel views for better image/feature matching, 
    local and global,  
    to improve both tracking and loop closing. 
    \item We propose a submap-based structure refinement following the global pose graph and point clouds adjustment to improve the reconstruction quality.
    \item We perform extensive experimental validation showing that 
    the proposed LGS-SLAM is able to improve the tracking accuracy by $70\%$ and the reconstruction quality by $50\%$ over the SOTA 3DGS-based SLAM methods.
\end{itemize}

%% file: 2_RelatedWork.tex
\section{Related Work}
\label{sec:related_work}

Traditional sparse/semi-dense visual SLAM systems can be divided into feature-based and direct methods. Feature-based methods~\cite{davison2007monoslam, klein2007parallel, geneva2020openvins, campos2021orb} focus on texture information in the scene and direct methods~\cite{engel2014lsd, forster2014svo, engel2017direct} are based on the grayscale invariant assumption.
Traditional dense visual SLAM systems~\cite{newcombe2011kinectfusion, dai2017bundlefusion} use RGB-D cameras to track and build dense maps. 
These methods only reliably account for the observed portions of the scene during the reconstruction. 




Implicit SLAM~\cite{sucar2021imap, zhu2022nice, sandstrom2023point} often uses NeRF to represent scenes. 
iMAP~\cite{sucar2021imap} uses a single MLP to represent the scene, demonstrating the feasibility of this approach. NICE-SLAM~\cite{zhu2022nice} and Co-SLAM~\cite{wang2023co} use a hierarchical feature grid with several MLP decoders to represent the map. Point-SLAM~\cite{sandstrom2023point} introduces a neural point cloud, further improving reconstruction accuracy. However, the rendering speed of NeRF remains a significant bottleneck and the implicit scene is not easy to edit.


The scene represented by 3D Gaussian points is easy to edit while facilitating faster rendering. Photo-SLAM~\cite{huang2024photo} maintains a hyper primitives map that stores ORB features~\cite{rublee2011orb} and Gaussian physical properties. The tracking and loop closure modules largely rely on ORB-SLAM3~\cite{campos2021orb} which uses only a motion-only bundle adjustment. Feature points often encounter repeatability of detection and mismatching errors, which cannot be eliminated during pose optimization. Moreover, the storage of ORB features for Gaussian points is not negligible, especially in large-scale scenarios. LoopSplat~\cite{zhu2024loopsplat} adds global BA to reduce accumulated errors and ensure overall spatial consistency. However, these methods are limited to indoor scenes.


%% file: 3_Method.tex
\section{LSG-SLAM}
\label{sec:method}

\begin{figure*}[t]
    \centering
    \includegraphics[width=0.85\linewidth]{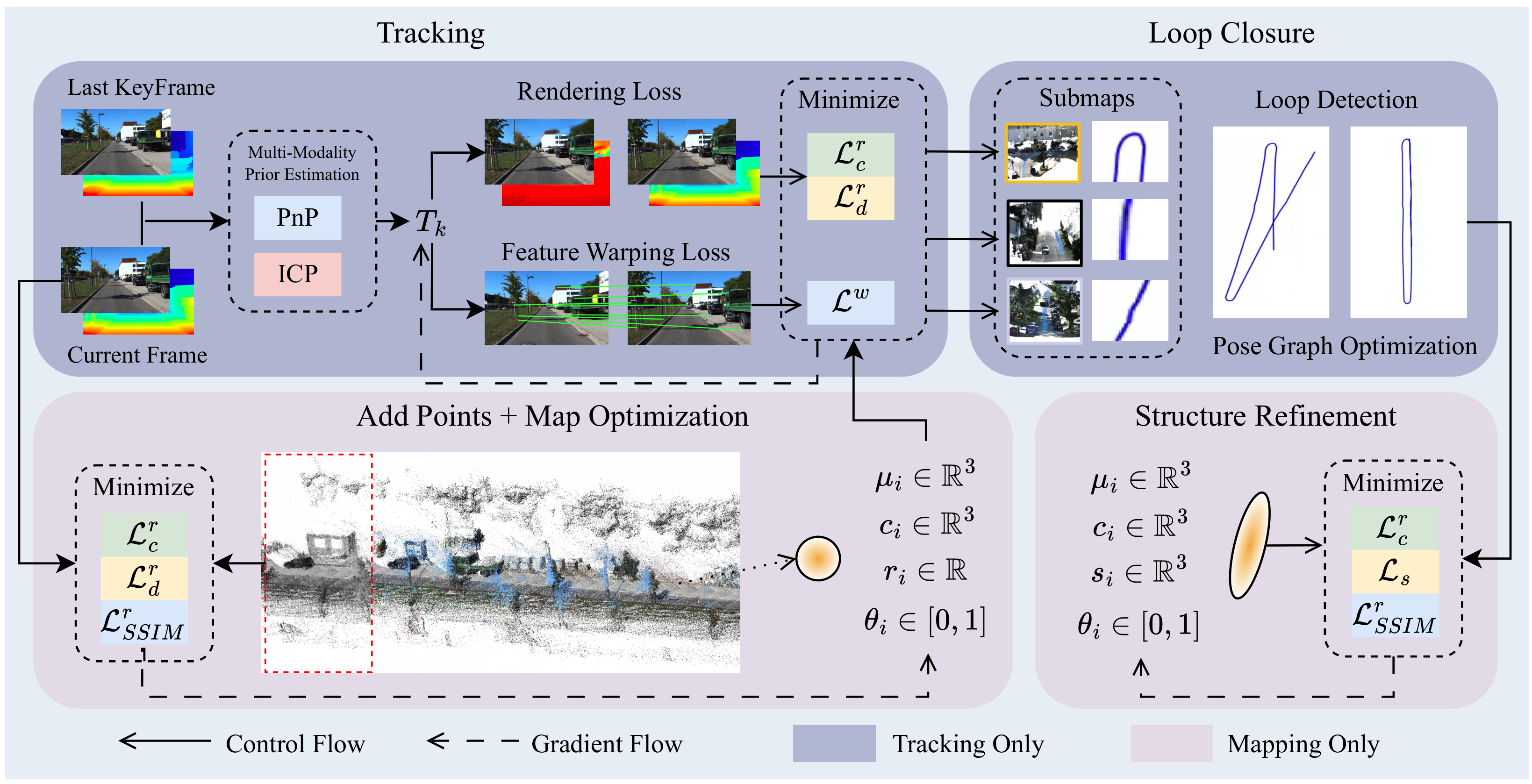}
    \caption{The overview of LSG-SLAM. For each incoming frame, we employ the multi-modality prior pose estimation strategy and subsequently utilize rendering loss and feature warping loss to optimize the current pose. Selected keyframes are then employed to refine the scene and add more Gaussian points. Incorporating loop detection and loop constraint estimation utilizing GS submaps, all keyframe poses are optimized. Subsequently, all Gaussian points are adjusted based on the relative transformation of their associated keyframes. After optimizing the pose of all images, a structure refinement process is executed to enhance the reconstruction quality for novel view synthesis.}
    \label{fig:pipeline}
    \vspace{-5mm}
\end{figure*}

The proposed LSG-SLAM is a stereo SLAM system that simultaneously tracks the camera pose and reconstructs the scene using 3D Gaussian points. 
Fig.~\ref{fig:pipeline} depicts the overall system architecture.
The primary components include simultaneous tracking and mapping of continuous GS submaps, loop closure, and structure refinement.
In the following, we will explain these modules in detail.



\subsection{3DGS Map Representation}
\label{3DGS_Map}
Gaussian points are the primary elements for 3DGS and are characterized by the following optical properties: RGB colors $\bm{c}$, opacity $o$, its center position $\bm{\mu}$, and the radius $r$. Following SplaTAM~\cite{keetha2024splatam}, we omit the spherical harmonics (SHs) for view-dependent radiance and enforce isotropy on Gaussian points. Encouraging sphericality avoids creating elongated Gaussian points along the viewing direction, which causes artifacts~\cite{matsuki2024gaussian}. 
Consequently, each Gaussian point is weighted by its opacity $o$ as follows:
\begin{equation}
\label{eq3}
f(\bm{x}) = o \exp(-\frac{\Vert\bm{x} - \bm{\mu}\Vert^2}{2r^2}).
\end{equation}
More details on splatting and rasterization can be referred to~\cite{kerbl20233d}. 
As in~\cite{keetha2024splatam}, we also render a silhouette image to ascertain the visibility of whether a pixel $\mathbf {p}=(u,v)$ contains information from the current map:
\begin{equation}
\label{eq:silhouette}
S(\mathbf {p}) = \sum_{i=1}^{n} f_i(\mathbf{p}) \prod_{j=1}^{i-1} (1 - f_j(\mathbf{p})).
\end{equation}
where $f(\mathbf{p})$is the splatted 2D Gaussian in pixel-space.

\subsection{Tracking}
%


Given a set of 3D Gaussian points and the camera pose, Gaussian Splatting can render a color image and a depth image through rasterization~\cite{kerbl20233d}. 
Rendering losses are constructed between the rendered images ($\mathbf{\widehat{I}}_\text{c}$, $\mathbf{\widehat{I}}_\text{d}$) and the actual images ($\mathbf{{I}}_\text{c}$, $\mathbf{{I}}_\text{d}$):
\begin{equation}
\label{eq:rendering_loss}
\begin{aligned}
    &\mathcal {L}^{r}_{c} = \sum_{\textbf {p}} \Bigl (S(\textbf {p}) > \theta \Bigr ) ({|| \mathbf{\widehat{I}}_\text{c} - \mathbf{I}_{c} ||_1}).
    \\
    &\mathcal {L}^{r}_{d} = \sum_{\textbf {p}} \Bigl (S(\textbf {p}) > \theta \Bigr ) ({|| \mathbf{\widehat{I}}_\text{d} - \mathbf{I}_{d} ||_1}).
    \\
    &{\mathcal{L}^{r}} =\lambda_{c}\mathcal {L}^{r}_{c} + \lambda_{d}\mathcal {L}^{r}_{d}.
\end{aligned}
\end{equation}
The camera pose is optimized through back propagation. 
Note that during this tracking, only the current image pose is optimized, while the 3DGS map is kept constant.
Since we consider stereo vision, the actual depth can be calculated from stereo disparity~\cite{xu2023iterative}. 
However, we find the rendering loss (which only maintains the consistency of appearance similarity) is fragile to large viewpoint changes, a common occurrence in large-scale scenarios,
and gradient descent-based optimization used in neural networks (back propagation) heavily depends on the initial guess and is easily trapped in poor local minima.


Therefore, instead of using a uniform motion model like MonoGS~\cite{matsuki2024gaussian} and SplaTAM~\cite{keetha2024splatam}, we design a multi-modality strategy for tracking with both 2D and 3D visual information. 
Specifically, we employ SuperPoint~\cite{detone2018superpoint} to extract local descriptors $\mathbf{LF}$ and LightGlue~\cite{lindenberger2023lightglue} for feature matching between the current image and the latest keyframe. The initial camera pose is then acquired through a PnP~\cite{nister2004efficient} solver with RANSAC~\cite{fischler1981random}. Moreover, if the number of 2D-3D inliers is below the threshold, we additionally employ the well-established point cloud registration~\cite{segal2009generalized} to obtain the initial pose from 3D. This simple yet effective method significantly enhances the reliability of the initial value, thereby reducing the difficulty of gradient descent optimization.

Moreover, the abundance of weak texture areas and large similar regions such as roads and sky can also impede the rendering losses. To further enhance the tracking accuracy, we incorporate a feature alignment scheme inspired by our previous work~\cite{xin2024hero}, explicitly supervising the spatial relationship between the keyframe and the current frame. Based on the 2D-2D correspondences established in the prior pose estimation module, given an inlier keypoint $i$ in the keyframe as $\mathbf{I}_{ki}$, its matched keypoint in the current image as $\mathbf{I}_{qj}$, the warping point of $\mathbf{{I}}_{ki}$ in the current image, named $\mathbf{I}_{qi'}$, can be calculated as:
\begin{equation}
    \mathbf{I}_{qi'} = \Pi(\mathbf{\widehat{D}}_{ki}\mathbf{R}^q_k\Pi^{-1}(\mathbf{\widehat{I}}_{ki}) + \mathbf{t}^q_k).
\end{equation}
where $\mathbf{\widehat{D}}_{ki}$ is the depth of $\mathbf{\widehat{I}}_{ki}$, $\Pi$ is the projection matrix of the intrinsic camera parameter and $\mathbf{R}^q_k$ and $\mathbf{t}^q_k$ are the relative rotation and translation from the keyframe to the current image.
The feature point and the feature map are supervised via pixel-wise loss functions:
\begin{equation}
\begin{aligned}
    &\mathcal{L}^w_\text{FP} = \frac{ \sum_{(i, j) \in S_{ij}} M_i ||\mathbf{I}_{qi'} - \mathbf{I}_{qj}||_2}{\sum_{ij} M_i}. &
    \\
    &\mathcal{L}^w_\text{FM} = \frac{\sum_{(i, j) \in S_{ij}} M_i \cdot ||\mathbf{LF}_j(\mathbf{I}_{qi'}) - \mathbf{LF}_j(\mathbf{I}_{qj})||_2}{\sum_{ij} M_i}.
    \\
    &{\mathcal{L}^{w}} =\mathcal{L}^w_\text{FP} + \gamma\mathcal{L}^w_\text{FM}.
\end{aligned}
\label{eq:feat_warp}
\end{equation}
where $\mathcal{L}^w_\text{FP}$ measures the pixel distance and $\mathcal{L}^w_\text{FM}$ measures the feature distance. 
Incorporating the feature map supervision is beneficial to mitigate the repeatability error of feature points. A mask $M$ is utilized to exclude keypoints with large depth, which are unstable in stereo depth estimation.


The overall  loss of tracking is a combination of rendering and warping losses:
\begin{equation}
\begin{aligned}
    \mathcal{L}_t = \mathcal{L}^{r} + \alpha\mathcal{L}^{w}.
\end{aligned}
\label{eq:total_loss}
\vspace{-2mm}
\end{equation}
Given that the rendering loss applies to all pixels of the entire image and the warping loss is only for the feature points, 
we have used a constant value $\alpha$ in Eq.\eqref{eq:total_loss} to maintain the proportion of the two losses.

\subsection{Mapping}


During mapping, the 3DGS parameters are optimized while keeping all camera poses fixed. 
Every few frames, we add the new frame to the keyframe list. Only keyframes participate in the mapping process and two categories of keyframes are selected. One is several keyframes that exhibit the highest overlap with the latest keyframe. The method to determine the degree of overlap is to convert the latest keyframe into a point cloud, project it onto the previous keyframes, and then count the number of points projected inside the image. The other consists of keyframes randomly selected from the keyframe set. The former is used to optimize the newly introduced Gaussian points, while the latter is employed to prevent the global map from being forgotten.


We incorporate the rendering loss similar to the tracking but without the use of the silhouette mask. Additionally, we add an SSIM~\cite{wang2004image} loss ${L}^{r}_{SSIM}$ for the color image, as the SSIM loss excels in preserving structural information and perceptual quality. The strategies for densification and culling of Gaussian points align with~\cite{kerbl20233d}. 
As such, the total mapping loss is given by:
\begin{equation}
\label{eq:ssim_loss}
\begin{aligned}
\mathcal{L}_m = \mathcal{L}^{r} +  \sum_{\textbf {p}} ({|| \mathbf{\widehat{I}}_\text{c} - \mathbf{I}_{c} ||_{SSIM} }) = \mathcal{L}^{r} + 
 {\mathcal {L}^{r}_{SSIM}}.
\end{aligned}
\end{equation}



Note that in large-scale mapping, reconstructing all frames into a comprehensive map is impractical due to the huge resource requirement of memory and GPUs. 
To address this issue,
we divide the entire scene into multiple continuous GS submaps based on the length of the trajectory. The last frame in the current submap is the same as the first frame in the subsequent submap. Each Gaussian point is attached to the keyframe that created it, facilitating the final global optimization. 


\subsection{Loop Closure}
To generate a globally consistent map, we incorporate loop closure to merge all GS submaps and eliminate accumulated errors. After the tracking and mapping of all submaps, we extract a global feature~\cite{wang2022transvpr} for each keyframe, and then employ place recognition~\cite{xin2019localizing} to find top $K$ similar reference keyframes in different GS submaps. 
Valid loops are then selected with enough inliers using point feature matching. 
Relative pose between looped keyframes can be found via relocalization based on the GS submap. 
Compared to tracking, loop closure involves more drastic changes in viewpoint and appearance. 
Our proposed multi-modality strategy and the integration of rendering and warping losses effectively alleviate these challenges.

The GS submap consists of a set of  Gaussian points and a keyframe database, where the database stores the global features, extracted 2D feature points, and poses of all keyframes. Note that no actual images are stored in the map. We render images on the fly for loop closure. 
Leveraging the fast rasterization of 3DGS, we can conveniently acquire color and depth images through rendering and construct the rendering loss with dense pixels, helping to mitigate errors from sparse 2D-3D point matching~\cite{campos2021orb}~\cite{qin2018vins}.
Moreover, we eliminate the storage of feature points in 3D like Photo-SLAM~\cite{huang2024photo}. Apart from reducing storage requirements, we demonstrate that the distribution of feature matching may differ in tracking and loop closure which involves appearance changes. As such, matching in 2D and then lifting to 3D with depth images is substantially more robust than direct 3D matching. 



Assuming we already have a valid loop $I_q$ in GS submap $\textbf{Q}$ and $I_r$ in $\textbf{R}$. For estimating the relative pose $T_{rq}$ between looped keyframes, we attempt to register $I_q$ into the submap $\textbf{R}$. We first render the color images $\widehat{^QI_q}$, $\widehat{^RI_r}$ and depth images $\widehat{^QD_q}$, $\widehat{^RD_r}$ using the keyframe poses $^{Q}T_q$ and $^{R}T_r$, then using the same multi-modality prior pose estimation in tracking, the initial value of $T^{'}_{rq}$ is estimated and $^{R}T^{'}_q = ^{R}T_r * T^{'}_{rq}$ is further obtained.
Next, we render color image $\widehat{^RI_q}$ and depth image $\widehat{^RD_q}$ and try to minimize the appearance difference and projection errors against ${^QI_q}$ and ${^QD_q}$. This is achieved by optimizing $^{R}T_q$ with rendering and warping losses. Finally, the loop constraint is calculated $T_{rq} = ^{R}T^{-1}_r * ^{R}T_q$.

After obtaining all valid loops and their relative poses, a keyframe-based pose graph is optimized, which includes adjacent and loop edges.
Subsequently, all Gaussian points are adjusted based on the relative transformation of their associated keyframes.

\subsection{GS Refinement}
Following Gaussian points adjustment, the structure refinement is performed to enhance the reconstruction quality of each submap. Specifically, we convert the radius $r_i$ of the Gaussian points to scale $s_i\in \mathbb{R}^3$, which represents the length of the Gaussian ellipsoid along the three coordinate axes. Converting each point from a sphere to an ellipsoid aids in representing the fine texture of the scene, and optimizing anisotropic 3D Gaussians can help complete reflective surfaces and skies.
Subsequently, the various attributes of the Gaussian points are optimized by rendering losses $\mathcal {L}_m$ against randomly sampled keyframes. In addition to rendering losses, the scale regularization loss~\cite{chen2023neusg} is employed to make each Gaussian ellipsoid close to flat, to better represent the surface of the object. 
For the scale attribute $s = (s_1, s_2, s_3) \in \mathbb{R}^3$ of the Gaussian ellipsoid, the scale regularization loss 
is $\mathcal{L}_s = || \min(s_1, s_2, s_3) ||_1$.


%% file: 4_Experiments.tex
\section{Experimental Results}
\label{sec:experiment}

In this section, we conduct extensive real-world experiments to validate the proposed LSG-SLAM.

\textbf{Datasets}. We use two well-known stereo datasets EuRoC and KITTI for evaluation. The EuRoC MAV dataset~\cite{burri2016euroc} combines indoor and outdoor scenes, with drastic viewpoint changes and massive illumination changes. The large-scale KITTI dataset~\cite{geiger2012we} encompasses diverse scenarios such as urban, rural, and highway.

\textbf{Metrics}. We evaluate LSG-SLAM in terms of both tracking and mapping. The root-mean-square error (RMSE) of the absolute trajectory error (ATE)~\cite{sturm2012benchmark} is utilized for trajectory accuracy. For reconstruction quality, we employ peak signal-to-noise ratio (PSNR), structural similarity index (SSIM)~\cite{wang2004image}, and learned perceptual image patch similarity (LPIPS)~\cite{zhang2018unreasonable}. 

\textbf{Baselines}. We benchmark LSG-SLAM against various representative SLAM techniques. These include traditional pipelines such as ORB-SLAM3~\cite{campos2021orb} and SVO~\cite{forster2014svo}, deep-learning-based systems like DROID-SLAM~\cite{teed2021droid} and PVO~\cite{ye2023pvo}, as well as recent 3DGS-based SLAM methods such as SplaTAM~\cite{keetha2024splatam}, MonoGS~\cite{matsuki2024gaussian}, and Photo-SLAM~\cite{huang2024photo}. For SplaTAM~\cite{keetha2024splatam} and MonoGS~\cite{matsuki2024gaussian}, we modify their depth acquisition technique to match ours, using stereo disparity, to ensure an equitable comparison. When evaluating the quality of reconstruction, the original 3D Gaussian Splatting~\cite{kerbl20233d} is included for comparison, its initial point cloud is generated with the ground truth poses and depth images. 

\textbf{Implementation Details}. To demonstrate that our method is robust to drastic viewpoint changes, we downsample the input image frequency for efficiency, setting the interval to 5 for EuRoC and 2 for KITTI. All comparison methods use full-frequency images. The tracking iteration is 50 for each frame and the mapping and loop optimization iteration is 100 for each keyframe. The Adam optimizer is employed for both camera pose and Gaussian parameters optimization. For stereo disparity, we utilize the pre-trained model of IGEV~\cite{xu2023iterative}. At most 512 SuperPoint~\cite{detone2018superpoint} keypoints are extracted per image. To balance different losses, we set $\lambda_c$ to 1.0, $\lambda_d$ to 0.2 in Eq.~\ref{eq:rendering_loss}, $\gamma$ to 10 in Eq.~\ref{eq:feat_warp} and $\alpha$ to 10 in Eq.~\ref{eq:total_loss} for all experiments. The threshold $\theta$ in the silhouette mask is 0.99 and the valid stereo depth threshold is 30 meters. These two masks are employed in calculating reconstruction quality metrics.




\input{tables/tab_track_euroc}

\input{tables/tab_mapping_euroc}

\subsection{Evaluation on EuRoC}
\textbf{Tracking}.
\label{sec:euroc-tracking}
Detailed comparison results are shown in Tab.~\ref{tab:track-euroc}. Our method significantly enhances tracking accuracy in comparison to 3DGS-based methods, even utilizing lower-frequency images. Methods like SplaTAM~\cite{keetha2024splatam} and MonoGS~\cite{matsuki2024gaussian} rely solely on the uniform motion model, which can easily cause tracking drift. Photo-SLAM~\cite{huang2024photo} minimizes reprojection errors of ORB features, yet indoor scenes with weak textures and drastic camera movements can cause the failure of feature matching. LSG-SLAM employs multi-modality prior estimation to handle drastic viewpoint changes and integrates both rendering losses and feature-alignment warping constraints for pose optimization. The former mitigates errors caused by non-repetitive feature point extraction and the impact of weak texture areas, while the latter reduces the misleading effects of appearance similarity in large similar regions. Following loop closure optimization, LSG-SLAM not only achieves comparable trajectory accuracy to ORB-SLAM3~\cite{campos2021orb} but also exhibits a higher reconstruction success rate in challenging scenarios.

\textbf{Mapping}.
Tab.~\ref{tab:mapping-euroc} presents the rendering results. Our method demonstrates superior rendering quality against SplaTAM~\cite{keetha2024splatam} and MonoGS~\cite{matsuki2024gaussian}, even in the absence of the structure refinement module. The enhanced tracking accuracy of our method also reduces the map structural errors. Upon incorporating the structure refinement module, the reconstruction quality has a significant enhancement, which demonstrates the ellipsoid is more effective than the sphere in capturing intricate texture details. Moreover, the added scale regularization loss results in a higher PSNR than the original 3D Gaussian Splatting~\cite{kerbl20233d}. 
For more qualitative results, please refer to our supplementary video.

\input{tables/tab_track_ablation}
\input{tables/tab_mapping_ablation}

\input{tables/tab_track_kitti}

\input{tables/tab_mapping_kitti}

\subsection{Evaluation on KITTI}
\textbf{Tracking}.
Due to a lack of consideration for memory consumption, representative 3DGS-based methods~\cite{keetha2024splatam, matsuki2024gaussian} cannot fully process the entire sequence. In contrast, through loop closure based on continuous GS submaps, LSG-SLAM is capable of reconstructing large-scale scenes with limited resources. The accuracy of pose estimation is compared in Tab.~\ref{tab:track-kitti}. Our method outperforms both classical and learning-based methods. Notably, methods like PVO~\cite{ye2023pvo} and DROID-SLAM~\cite{teed2021droid} require several days for network training, and the generalization capability is limited. In contrast, our method does not necessitate training and is scene-independent.

\textbf{Mapping}.
Tab.~\ref{tab:mapping-kitti} shows that our structure refinement module significantly improves the rendering quality. 3D Gaussian Splatting~\cite{kerbl20233d} directly optimizes anisotropic Gaussian ellipsoids, which often leads to floaters. In contrast, isotropic Gaussian spheres converge faster and are less prone to floaters in the early optimization stages. Our method first reconstructs the scene using isotropic Gaussian spheres, learning a good initial value. Then, in the structure refinement stage, the spheres are transformed into ellipsoids to refine the surface details of the objects. This procedure enhances the resilience of our method to floaters, leading to improved rendering quality. For more qualitative results, please refer to our supplementary video.

\subsection{Ablation Study}
\textbf{Tracking}.
We select several sample submaps from the KITTI dataset to test the impact of the submap length and different loss combinations, as shown in Tab.~\ref{tab:track-ablation}.
The length of GS submaps does not significantly influence the localization accuracy. Rendering losses are highly dependent on the quality of the prior pose.
However, solely relying on prior estimation cannot yield reliable results. 
When only using feature point warping loss, the ATE RMSE of SubSeq 08 drops from 1.01m to 1.06m, which demonstrates the effectiveness of the feature map warping loss. This is because detection and matching processes often encounter repeatability errors in feature points, which cannot be eliminated during pose optimization in tracking. Incorporating feature-metric supervision to maintain the semantic information’s consistency around feature points is crucial for reducing errors.

\textbf{Mapping}.
We selected two sequences from the KITTI dataset to test the effects of the shape of Gaussian points and the effectiveness of the scale regularization loss, as shown in Tab.~\ref{tab:mapping-ablation}. Changing the Gaussian point from a sphere to an ellipsoid improves the rendering quality significantly, as the ellipsoid can become elongated after further optimization to represent the detailed texture in the scene. The scale regularization loss forces the ellipsoid to become flatter, making each Gaussian point better represent the surface of the object.

\subsection{Efficiency Analysis}

In Tab.~\ref{tab:efficiency}, we analyze the elapsed time of each stage in tracking and mapping. The experiments are conducted on a desktop PC with a 3.60GHz Intel Core i9-9900K CPU and an NVIDIA RTX 3090 GPU. Given the robustness to viewpoint changes and the ability to operate at low frequencies, LSG-SLAM showcases its potential to operate in real-time. The speedup can be achieved by employing multithread processing and the early stop strategy of monitoring the reduction of losses.

\input{tables/tab_efficiency}

%% file: tables/tab_track_euroc.tex
\begin{table}[t]
\setlength{\tabcolsep}{0.95mm}
\renewcommand\arraystretch{1.1}
\caption{Camera tracking results on EuRoC (ATE RMSE $\downarrow$[m]). "\ding{56}" means tracking lost, "-" denotes results not publicly available.}
\vspace{-2mm} 
\label{tab:track-euroc}
\hspace*{0cm}\makebox[\linewidth][c]{%
\begin{tabular}{ c  c c c c c  c c c c }
\toprule
{Method (loop)} & MH02 & MH03 & V101 & V102 & V201 & V203 & Avg. \\
\hline
SVO~\cite{forster2014svo} (w/o)& 0.12 & 0.41 & \textbf{0.07} & 0.21 & 0.11 & 1.08 & {\bf \color{silver} 0.33}\\
Splatam~\cite{keetha2024splatam} (w/o)& 0.39 & 4.78 & 0.09 & 0.26 & 0.27 & 3.85 & 1.61\\
MonoGS~\cite{matsuki2024gaussian} (w/o) & \textbf{0.06} & 1.81 & 0.10 & 0.17 & 0.58 & 1.89 & 0.77\\
Ours (w/o) & \textbf{0.06} & \textbf{0.15} & 0.12 & \textbf{0.07} & \textbf{0.09} & \textbf{0.53} & {\bf \color{gold} 0.17}\\
\midrule
ORB-SLAM3~\cite{campos2021orb} (w) & 0.03 & \textbf{0.03} & \textbf{0.03} & \textbf{0.02} & \textbf{0.02} & \ding{56} & {\bf \color{gold} 0.03}\\
Photo-SLAM~\cite{huang2024photo} (w) & 0.04 & - & 0.09 & - & 0.27 & - & 0.13\\
Ours (w) & \textbf{0.02} & 0.08 & 0.06 & 0.04 & 0.05 & \textbf{0.13} & {\bf \color{silver} 0.06}\\
\bottomrule
\end{tabular}}
\vspace{-3mm}
\end{table}

%% file: tables/tab_mapping_euroc.tex
\begin{table}[ht]
\setlength{\tabcolsep}{0.8mm}
\renewcommand\arraystretch{1.1}
\caption{Rendering results on EuRoC. "cr" represents the color refine in MonoGS, and "sr" represents the structure refinement in our method.}
\vspace{-2mm} 
\label{tab:mapping-euroc}
\hspace*{0cm}\makebox[\linewidth][c]{
\begin{tabular}{ccccccccc}
\toprule
Method                          & \multicolumn{1}{l}{Metrics} & \multicolumn{1}{l}{MH02} & \multicolumn{1}{l}{MH03} & \multicolumn{1}{l}{V101} & \multicolumn{1}{l}{V102} & \multicolumn{1}{l}{V201} & \multicolumn{1}{l}{V203} & \multicolumn{1}{l}{Avg.} \\ \hline
\multirow{3}{*}{3DGS~\cite{kerbl20233d}}           & PSNR$\uparrow$                        & 30.30                    & \textbf{30.37}                    & 28.44                    & \textbf{30.98}                    & 31.96                    & \textbf{32.85}                    & \color{silver} 30.81                    \\  
                                & SSIM$\uparrow$                        & 0.98                     & \textbf{0.98}                     & 0.97                     & \textbf{0.98}                     & \textbf{0.99}                     & \textbf{0.98}                     & \color{gold} 0.98                     \\ 
                                & LPIPS$\downarrow$                       & 0.04                     & \textbf{0.04}                     & \textbf{0.05}                     & \textbf{0.03}                     & \textbf{0.02}                     & \textbf{0.04}                     & \color{gold} 0.03                     \\ \hline
\multirow{3}{*}{Splatam~\cite{keetha2024splatam}}        & PSNR$\uparrow$                        & 17.84                    & 13.92                    & 21.16                    & 18.65                    & 18.08                    & 7.69                     & 16.22                    \\  
                                & SSIM$\uparrow$                        & 0.67                     & 0.53                     & 0.86                     & 0.73                     & 0.72                     & 0.18                     & 0.61                     \\  
                                & LPIPS$\downarrow$                       & 0.41                     & 0.49                     & 0.29                     & 0.40                     & 0.41                     & 0.79                     & 0.46                     \\ \hline
\multirow{3}{*}{\shortstack{MonoGS~\cite{matsuki2024gaussian} \\ (w/o cr)}} & PSNR$\uparrow$                        & 16.59                    & 13.21                    & 17.61                    & 19.24                    & 17.41                    & 14.91                    & 16.50                    \\
                                & SSIM$\uparrow$                        & 0.66                     & 0.47                     & 0.72                     & 0.75                     & 0.68                     & 0.63                     & 0.65                     \\ 
                                & LPIPS$\downarrow$                       & 0.29                     & 0.57                     & 0.42                     & 0.41                     & 0.44                     & 0.62                     & 0.46                     \\ \hline
\multirow{3}{*}{\shortstack{Ours \\ (w/o sr)}}   & PSNR$\uparrow$                        & 27.13                    & 23.36                    & 27.13                    & 24.41                    & 28.98                    & 25.14                    & 26.02                    \\ 
                                & SSIM$\uparrow$                        & 0.96                     & 0.93                     & 0.97                     & 0.94                     & 0.97                     & 0.94                     & \color{silver} 0.95                     \\ 
                                & LPIPS$\downarrow$                       & 0.07                     & 0.10                     & 0.06                     & 0.08                     & 0.05                     & 0.08                     & 0.07                     \\ \hline
\multirow{3}{*}{\shortstack{MonoGS~\cite{matsuki2024gaussian} \\ (w cr)}}  & PSNR$\uparrow$                        & 27.05                    & 18.03                    & 24.09                    & 24.00                    & 21.78                    & 19.18                    & 22.36                    \\ 
                                & SSIM$\uparrow$                        & 0.88                     & 0.66                     & 0.85                     & 0.83                     & 0.81                     & 0.78                     & 0.80                     \\ 
                                & LPIPS$\downarrow$                       & 0.13                     & 0.40                     & 0.24                     & 0.29                     & 0.28                     & 0.46                     & 0.30                     \\ \hline
\multirow{3}{*}{\shortstack{Ours \\ (w sr)}}    & PSNR$\uparrow$                        & \textbf{33.31}                    & 30.26                    & \textbf{31.31}                    & 30.91                    & \textbf{32.70}                    & 29.81                    & \color{gold} 31.38                    \\ 
                                & SSIM$\uparrow$                        & \textbf{0.99}                     & \textbf{0.98}                     & \textbf{0.98}                     & 0.97                     & \textbf{0.99}                     & 0.96                     & \color{gold} 0.98                     \\
                                & LPIPS$\downarrow$                       & \textbf{0.03}                     & 0.05                     & \textbf{0.05}                     & 0.05                     & 0.04                     & 0.12                     & \color{silver} 0.05                     \\ 
\bottomrule
\end{tabular}}
\vspace{-3mm}
\end{table}

%% file: tables/tab_track_ablation.tex
\begin{table}[ht]
\setlength{\tabcolsep}{1.3mm}
\renewcommand\arraystretch{1.1}
\caption{Tracking ablation Study on KITTI dataset. 
} 
\vspace{-1mm}
\label{tab:track-ablation}
\hspace*{0cm}\makebox[\linewidth][c]{%
\begin{tabular}{ c | c | c c | c c }
\toprule
Trajectory & Prior & \multicolumn{2}{c | }{Losses} & \multicolumn{2}{c}{ATE RMSE $\downarrow$[m]} \\
Length $[$m$]$ & Pose & Rendering & Warping & SubSeq 00 & SubSeq 08 \\ 
\hline
200 & \ding{56} & \ding{52} & \ding{56} & 6.17 & 62.21 \\
200 & \ding{52} & \ding{56} & \ding{56} & 1.42 & 2.02 \\
200 & \ding{52} & \ding{56} & \ding{52} & 0.56 & 1.30 \\
200 & \ding{52} & \ding{52} & \ding{56} & 0.48 & 1.15 \\
200 & \ding{52} & \ding{52} & \ding{52} & 0.47 & 1.01 \\
100 & \ding{52} & \ding{52} & \ding{52} & 0.48 & 0.97 \\
\bottomrule
\end{tabular}}
\vspace{-5mm}
\end{table}

%% file: tables/tab_mapping_ablation.tex
\begin{table}[ht]
\setlength{\tabcolsep}{2.0mm}
\renewcommand\arraystretch{1.1}
\caption{Rendering ablation study on KITTI dataset. 
}
\vspace{-2mm} 
\label{tab:mapping-ablation}
\hspace*{0cm}\makebox[\linewidth][c]{
\begin{tabular}{c|c|c c c}
\toprule
Ellipsoid          & Scale Regularization Loss & Metrics & 03    & 09    \\ \hline
\multirow{3}{*}{\ding{56}} & \multirow{3}{*}{\ding{56}}        & PSNR$\uparrow$    & 20.51 & 20.91 \\  
                   &                           & SSIM$\uparrow$    & 0.85  & 0.86  \\  
                   &                           & LPIPS$\downarrow$   & 0.21  & 0.20  \\ \hline
\multirow{3}{*}{\ding{52}} & \multirow{3}{*}{\ding{56}}        & PSNR$\uparrow$    & 24.89 & 25.38 \\  
                   &                           & SSIM$\uparrow$    & 0.95  & 0.95  \\  
                   &                           & LPIPS$\downarrow$   & 0.10  & 0.09  \\ \hline
\multirow{3}{*}{\ding{52}} & \multirow{3}{*}{\ding{52}}        & PSNR$\uparrow$    & 26.18 & 26.81 \\  
                   &                           & SSIM$\uparrow$    & 0.97  & 0.97  \\  
                   &                           & LPIPS$\downarrow$   & 0.07  & 0.07  \\
\bottomrule
\end{tabular}}
\vspace{-5mm}
\end{table}

%% file: tables/tab_track_kitti.tex
\begin{table*}[ht]
\setlength{\tabcolsep}{2.4mm}
\renewcommand\arraystretch{1.1}
\caption{Camera tracking results on KITTI (ATE RMSE $\downarrow$[m]). The results of sequence 01 are not reported, as all methods perform poorly on it.} 
\vspace{-2mm} 
\label{tab:track-kitti}
\hspace*{0cm}\makebox[\linewidth][c]{%
\begin{tabular}{c c c c c c c c c c c c }
\toprule
Method (Loop) & 00 & 02 & 03 & 04 & 05 & 06 & 07 & 08 & 09 & 10 & Avg. \\
\hline
DSO~\cite{engel2017direct} (w/o) & 113.18 & 116.81 & 1.39 & \textbf{0.42} & 47.46 & 55.62 & 16.72 & 111.08 & 52.23 & 11.09 & 52.60 \\
ORB-SLAM3~\cite{campos2021orb} (w/o) & 40.65 & 47.82 & 0.94 & 1.30 & 29.95 & 40.82 & 16.04 & 43.09 & 38.77 & 5.42 & 26.48 \\
PVO~\cite{ye2023pvo} (w/o) & 5.69 & 23.60 & \textbf{0.86} & 0.81 & 8.41 & 13.57 & 8.89 & \textbf{6.67} & 14.65 & 8.66 & {\bf \color{gold} 9.18} \\
DynaSLAM~\cite{bescos2018dynaslam} (w/o) & 8.07 & 21.78 & 0.87 & 1.40 & \textbf{4.46} & 14.36 & \textbf{2.63} & 50.37 & 41.91 & 7.52 & {\bf \color{bronze} 15.34} \\
DROID-SLAM~\cite{teed2021droid} (w/o) & \textbf{4.86} & 18.81 & 0.89 & 0.82 & 16.03 & 42.79 & 27.40 & 16.34 & 46.40 & 11.31 & 18.56 \\
SC-SfMLearner~\cite{bian2019unsupervised} (w/o) & 93.04 & 70.37 & 10.21 & 2.97 & 40.56 & \textbf{12.56} & 21.01 & 56.15 & 15.02 & 20.19 & 34.21 \\
Ours (w/o) & 24.06 & \textbf{13.39} & 1.88 & 1.62 & 13.39 & 19.82 & 3.36 & 8.59 & \textbf{11.88} & \textbf{2.06} & {\bf \color{silver}10.01} \\
\midrule
ORB-SLAM3~\cite{campos2021orb} (w) & 6.03 & 14.76 & \textbf{1.02} & \textbf{1.57} & 4.04 & 11.16 & 2.19 & 38.85 & 8.39 & 6.63 & {\bf \color{silver} 9.46} \\
Ours (w) & \textbf{3.15} & \textbf{9.22} & 1.88 & 1.62 & \textbf{1.97} & \textbf{2.62} & \textbf{1.52} & \textbf{8.43} & \textbf{6.06} & \textbf{1.98} & {\bf \color{gold} 3.85} \\

\bottomrule
\end{tabular}}
\vspace{-2mm}
\end{table*}

%% file: tables/tab_mapping_kitti.tex
\begin{table*}[]
\setlength{\tabcolsep}{2.4mm}
\renewcommand\arraystretch{1.1}
\caption{Rendering results on KITTI. "sr" represents the structure refinement module in our method.}
\vspace{-2mm} 
\label{tab:mapping-kitti}
\hspace*{0cm}\makebox[\linewidth][c]{
\begin{tabular}{cclllllllllll}
\toprule
Method                        & \multicolumn{1}{l}{Metrics} & 00    & 02    & 03    & 04    & 05    & 06    & 07    & 08    & 09    & 10    & Avg.  \\ \hline
\multirow{3}{*}{3DGS~\cite{kerbl20233d}}         & PSNR$\uparrow$                        & 23.62 & 23.08 & \textbf{26.28} & \textbf{26.99} & 25.59 & \textbf{28.31} & 25.64 & 23.06 & 23.64 & 25.30 & \color{silver}25.15 \\ 
                              & SSIM$\uparrow$                        & 0.92  & 0.91  & 0.95  & \textbf{0.96}  & 0.95  & \textbf{0.97}  & 0.94  & 0.91  & 0.92  & 0.94  & \color{silver}0.94  \\  
                              & LPIPS$\downarrow$                       & 0.15  & 0.18  & 0.10  & 0.10  & 0.11  & 0.07  & 0.10  & 0.17  & 0.17  & 0.12  & \color{silver}0.13  \\ \hline
\multirow{3}{*}{Ours (w/o sr)} & PSNR$\uparrow$                        & 21.03 & 21.06 & 20.51 & 22.71 & 20.85 & 22.74 & 20.27 & 20.24 & 20.91 & 20.66 & 21.10 \\  
                              & SSIM$\uparrow$                        & 0.88  & 0.87  & 0.85  & 0.89  & 0.86  & 0.90  & 0.86  & 0.85  & 0.86  & 0.85  & 0.87  \\ 
                              & LPIPS$\downarrow$                       & 0.18  & 0.20  & 0.21  & 0.15  & 0.20  & 0.15  & 0.19  & 0.21  & 0.20  & 0.21  & 0.19  \\ \hline
\multirow{3}{*}{Ours (w/ sr)}  & PSNR$\uparrow$                        & \textbf{27.09} & \textbf{26.64} & 26.18 & 25.46 & \textbf{26.90} & 27.79 & \textbf{25.98} & \textbf{26.17} & \textbf{26.81} & \textbf{26.82} & \color{gold}26.58 \\ 
                              & SSIM$\uparrow$                        & \textbf{0.97}  & \textbf{0.97}  & \textbf{0.97}  & 0.95  & \textbf{0.97}  & \textbf{0.97}  & \textbf{0.96}  & \textbf{0.96}  & \textbf{0.97}  & \textbf{0.97}  & \color{gold}0.97  \\ 
                              & LPIPS$\downarrow$                       & \textbf{0.07}  & \textbf{0.07}  & \textbf{0.07}  & \textbf{0.09}  & \textbf{0.07}  & \textbf{0.06}  & \textbf{0.07}  & \textbf{0.08}  & \textbf{0.07}  & \textbf{0.07}  & \color{gold}0.07  \\
\bottomrule
\end{tabular}}
\vspace{-4mm}
\end{table*}

%% file: tables/tab_efficiency.tex
\begin{table}[ht]
\setlength{\tabcolsep}{2.4mm}
\caption{Elapsed time of each stage in tracking and mapping. } 
\label{tab:efficiency}
\hspace*{0cm}\makebox[\linewidth][c]{%
\begin{tabular}{ r l }
\toprule
Stage & Elapsed time\\
\hline
Stereo Depth Estimation & 90ms / image \\
Global Feature Extraction & 16ms / image \\
SuperPoint Extraction & 13ms / image \\
LightGlue Matching & 35ms / image pair \\
Pose Prior Estimation & 100ms / image \\
Opt by 3DGS & 15ms / iter \\
Place Recognition & 0.01ms / image pair \\

\bottomrule
\end{tabular}}
\vspace{-3mm}
\end{table}

%% file: 5_Conclusion.tex
\section{Conclusions and Future Work}

This paper presents LSG-SLAM, the first of its kind 3DGS-based visual SLAM system specifically designed for large-scale scenarios only using stereo cameras. The primary components include simultaneous tracking and mapping of continuous submaps, loop closure, and structure refinement.
The proposed LSG-SLAM significantly enhances tracking stability, mapping consistency, scalability, and reconstruction quality. 
While our LSG-SLAM achieves state-of-the-art performance against both classical and learning-based methods, we did find that our method shows sensitivity to moving objects. 
A possible solution would be incorporating more semantic information~\cite{yan2024street} and we plan to explore this idea in the future work.
